# Artificial Intelligence-Based Multiscale Temporal Modeling for Anomaly Detection in Cloud Services

Lian Lian
University of Southern California
Los Angeles, USA

Yilin Li
Carnegie Mellon University
Pittsburgh, USA

Song Han
Northeastern University
Boston, USA

Renzi Meng
Northeastern University
Boston, USA

Sibo Wang
Rice University
Houston, USA

Ming Wang*
Trine University
Phoenix, USA

*Abstract-This study proposes an anomaly detection method based on the Transformer architecture with integrated multiscale feature perception, aiming to address the limitations of temporal modeling and scale-aware feature representation in cloud service environments. The method first employs an improved Transformer module to perform temporal modeling on high-dimensional monitoring data, using a self-attention mechanism to capture long-range dependencies and contextual semantics. Then, a multiscale feature construction path is introduced to extract temporal features at different granularities through downsampling and parallel encoding. An attention-weighted fusion module is designed to dynamically adjust the contribution of each scale to the final decision, enhancing the model's robustness in anomaly pattern modeling. In the input modeling stage, standardized multidimensional time series are constructed, covering core signals such as CPU utilization, memory usage, and task scheduling states, while positional encoding is used to strengthen the model's temporal awareness. A systematic experimental setup is designed to evaluate performance, including comparative experiments and hyperparameter sensitivity analysis, focusing on the impact of optimizers, learning rates, anomaly ratios, and noise levels. Experimental results show that the proposed method outperforms mainstream baseline models in key metrics, including precision, recall, AUC, and F1-score, and maintains strong stability and detection performance under various perturbation conditions, demonstrating its superior capability in complex cloud environments.*

## I. INTRODUCTION

With the rapid proliferation of cloud computing, cloud services have become the core components supporting modern information infrastructure. However, as cloud platforms continue to grow in scale and complexity, their operating environments exhibit high dynamism, multi-tenancy, and parallel multitasking[1]. These characteristics significantly increase uncertainty in system operations, making anomaly detection a critical task to ensure the stability and security of cloud services. Anomalies in cloud systems appear in diverse forms, including sudden workload spikes, response latency abnormalities, resource scheduling chaos, and container failures. If not detected and addressed promptly, such anomalies can lead to large-scale service interruptions, resource waste, and security breaches. Therefore, developing efficient, accurate, and scalable anomaly detection algorithms is of great theoretical and practical significance for improving platform performance, enhancing service robustness, and ensuring fault tolerance[2].

Traditional anomaly detection methods in the traditional domain often rely on rule matching, statistical thresholds, or supervised learning models [3-6]. These approaches perform well with static structures or low-dimensional data [7-10]. However, they usually fail when applied to massive, high-dimensional, heterogeneous, and non-stationary monitoring data from cloud services. This is mainly due to their limited generalization capabilities and poor sensitivity to temporal dynamics. In real-world scenarios, anomalies are typically sparse, with incomplete or even missing labels, which further challenges model trainability and generalization. Moreover, monitoring data collected from cloud platforms is highly multimodal and multiscale. For example, metrics such as CPU utilization, memory usage, service call chains, and network logs involve complex temporal dependencies and semantic couplings. Traditional models struggle to capture such latent relationships, resulting in low detection accuracy and response efficiency.

These limitations have been repeatedly observed in real systems. For example, in a large online service, Liu et al. report that selecting and tuning thresholds for a diverse set of statistical detectors is labor-intensive, and teams often end up "settling with static threshold-based detection" that performs poorly under non-stationary workloads; their Opprentice system had to combine 14 detectors and use supervised learning to meet operators' precision/recall preferences, underscoring the brittleness of fixed thresholds at scale [11]. Likewise, Twitter engineers observed that many off-the-shelf anomaly detectors are not applicable to cloud metrics because strong seasonality and trend violate distributional assumptions; they introduced the Seasonal Hybrid ESD method to explicitly handle these effects [12]. Benchmarking efforts such as the Numenta Anomaly Benchmark further stress the need for detectors that adapt online and avoid false positives on real-world streaming time series [13]. These experiences motivate models that capture long-range dependencies and multi-scale patterns rather than relying on static rules or single-scale temporal models. Therefore, designing models capable of perceiving temporal dynamics, multiscale features, and structural hierarchies has become a key research direction.

In recent years, the Transformer architecture has achieved remarkable success in fields such as natural language processing and computer vision [14-18]. Its strong sequence modeling capabilities and ability to capture global dependencies have been widely recognized. The self-attention mechanism enables flexible modeling of dependencies between any positions without predefined windows or fixed rules. This alleviates the limitations of traditional models in handling temporal context and multidimensional signal fusion. Applying Transformers to cloud service anomaly detection is expected to overcome existing bottlenecks in temporal modeling and semantic interaction. It enables fine-grained perception of complex anomaly patterns[19]. However, the standard Transformer architecture often suffers from high computational complexity and limited ability to model local patterns when processing high-frequency, long-span, and multimodal cloud service data. These issues constrain its scalability and practical deployment. Therefore, integrating multiscale fusion mechanisms into the Transformer while preserving its global modeling ability is essential for performance and adaptability[20].

Multiscale fusion mechanisms allow the extraction of critical features across different time granularities, feature spaces, and receptive fields [21-23]. This enhances the model's ability to perceive both local fluctuations and global trends. In cloud service monitoring, behavioral changes at different time scales often contain distinct anomaly patterns[24]. For instance, microsecond-level jitter and minute-level resource leakage exhibit different feature distributions and semantic characteristics. Relying on a single-scale model can easily miss key anomaly signals. Deeply integrating multiscale fusion with the Transformer architecture enhances the capacity to capture multilevel temporal features and identify multi-granularity anomaly patterns, which is particularly important in supporting the intensive computational demands of large language models in the current AI era [25-28].

In practical cloud service operations, anomaly detection systems must ensure high accuracy and low false positive rates. At the same time, they need to meet requirements for real-time response and resource-efficient deployment. These impose strict constraints on algorithm design and computational efficiency. Introducing lightweight Transformer-based structures and multiscale attention mechanisms can significantly reduce computational costs while maintaining expressive power. This improves deployment efficiency and runtime stability. As a result, intelligent perception and dynamic monitoring of complex behaviors in large-scale cloud environments become feasible. This study focuses on combining Transformer architectures with multiscale feature fusion mechanisms. It aims to explore efficient anomaly detection strategies for cloud service environments and provide a theoretical and technical foundation for the next generation of intelligent operations and maintenance systems.

## II. RELATED WORK

Anomaly detection in cloud and distributed environments has long been a critical research area, driven by the increasing complexity and scale of cloud platforms. Early work in this field focused on improving the reliability of clustered cloud systems by employing statistical and machine learning-based anomaly detection tailored to high-dimensional and dynamic architectures [29]. These approaches demonstrated the importance of addressing system-specific patterns and highlighted the limitations of conventional rule-based strategies. With the emergence of microservice architectures, specialized models such as those utilizing conditional multiscale generative adversarial networks (GANs) and adaptive temporal autoencoders have been proposed to address the highly modular and fine-grained anomaly patterns seen in microservice ecosystems [30]. Such frameworks offer strong capabilities in extracting subtle anomalies that often elude traditional detectors.

In parallel, deep learning methods have begun to dominate anomaly detection research in cloud systems due to their superior feature extraction and modeling capabilities. Self-attention-based approaches allow for dynamic modeling of multi-source cloud metrics, improving both trend prediction and anomaly identification by effectively capturing dependencies across heterogeneous signals [31]. Further, graph-based attention optimization has enabled cloud systems to detect malicious user patterns with greater precision, leveraging the structural relationships inherent in user interactions and system events [32]. The integration of multi-modal information, supported by deep learning for root cause detection, has further expanded the potential for comprehensive anomaly diagnostics, offering better interpretability and actionable insights in distributed environments [33].

The introduction of Transformer architectures has marked a turning point for time series anomaly detection. Innovations such as patch-based Transformers, which emphasize localized reconstruction errors, provide improved interpretability and sensitivity to localized temporal anomalies—key for complex monitoring scenarios [34]. Additionally, deep learning frameworks that incorporate bidirectional LSTMs and multi-scale attention, as well as multi-scale feature reconstruction networks, have significantly advanced the granularity and robustness of anomaly detection, particularly in industrial and multi-domain contexts [35-36]. Temporal fusion Transformers have been successfully applied to multi-horizon time series forecasting, facilitating interpretable and context-aware temporal modeling across various application domains [37].

A further methodological advance lies in the integration of multiscale feature extraction, graph convolution, and multi-modal learning architectures. For example, joint graph convolution and sequential modeling have proven effective in handling the scalability and complexity of network traffic estimation, demonstrating the utility of graph-based learning for cloud-scale monitoring [38]. Time-aware generative diffusion frameworks have enabled accurate modeling of volatility in high-frequency temporal data, introducing new paradigms for uncertainty quantification and predictive analytics [39]. While spatiotemporal feature learning networks were originally developed for human activity recognition, their ability to extract hierarchical multiscale features is highly relevant for understanding intricate anomaly patterns in cloud environments [40].

Adaptive resource management in large-scale distributed and cloud systems has also benefited from reinforcement learning and federated learning techniques. Deep Q-Networks (DQNs) and edge-based coordination approaches have been leveraged for dynamic operating system scheduling and IoT scheduling, enabling systems to self-optimize in real time by learning from the operational environment [41-42]. Autonomous resource management in microservices through reinforcement learning has further improved system efficiency and adaptability, meeting the growing demands of highly dynamic service topologies [43]. Federated learning, both in its standard and privacy-enhanced forms, has addressed the need for secure, collaborative model training across distributed and heterogeneous data sources, supporting robust and privacy-preserving operations in cloud environments [44-45].

Feature extraction and dimensionality reduction remain central to effective high-dimensional anomaly detection. Autoencoder-based data mining frameworks have proven valuable for extracting salient features and reducing complexity in massive monitoring datasets, thus facilitating downstream anomaly detection and prediction tasks [46]. Additionally, dynamic Transformer-based rule mining has contributed to improved interpretability and adaptability, allowing for the discovery of contextual patterns and the generation of actionable anomaly explanations in complex environments [47]. Finally, the use of advanced load forecasting and meta-learned representations demonstrates the importance of transfer learning and meta-learning techniques for generalizing anomaly detection models across varying cloud workloads [48]. These approaches offer robust adaptation to new tasks and conditions, supporting resilient anomaly detection in continuously evolving cloud ecosystems.

Collectively, these diverse methodologies form the foundation for the present work, which synthesizes Transformer-based architectures with multiscale feature perception and attention-weighted fusion. By building on innovations in self-attention, multi-scale modeling, and adaptive learning, the proposed method aims to address the persistent challenges of temporal modeling, hierarchical feature extraction, and robust anomaly detection in complex, real-world cloud service environments.

### III. METHOD

#### A. Transformer Architecture

In this study, we adopt an improved Transformer architecture as the foundational backbone for our anomaly detection framework, aiming to fully exploit the temporal dependencies and global feature correlations present in the multi-dimensional monitoring data of cloud services. The Transformer, originally designed for natural language processing tasks, is known for its powerful self-attention mechanism, which enables comprehensive modeling of pairwise relationships across all input elements. Leveraging this characteristic, we extend its applicability to the time series domain, where long-range dependencies and subtle temporal patterns are critical for accurate anomaly identification.

To tailor the Transformer to the unique challenges of cloud service monitoring, we preserve its core structure while introducing targeted adaptations at the input level. These enhancements include the design of standardized temporal input representations, a refined position encoding scheme that captures time-sensitive information more effectively, and a customized attention allocation strategy to emphasize critical patterns associated with abnormal behaviors. These modifications are carefully integrated to improve the model's ability to capture both global trends and localized anomalies in complex, dynamic environments. The complete model architecture is illustrated in Figure 1.

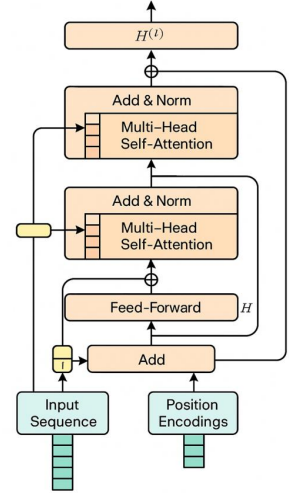

Figure 1. Transformer Model Architecture

First, the input data is represented as a sequence of feature vectors at consecutive time steps, defined as:

$$X = [x_1, x_2, ..., x_T], x_t \in R^d$$

Among them, $T$ represents the sequence length, $d$ represents the feature dimension of each time step, and $x_t$ is the observed feature vector at the *t-th* time point.

To preserve the timing information, we introduce a learnable position encoding vector $P = [p_1, p_2, ..., p_T]$ and add it to the input sequence position by position to obtain the position-aware input sequence:

$$H^{(0)} = X + P$$

The embedding is then processed in a multi-layer Transformer encoder. Each layer of the encoder consists of a multi-head self-attention module and a feed-forward neural network with residual connections and layer normalization operations. In layer l, the query, key, and value are generated by linear transformations, respectively, as defined below:

$$Q^{(l)} = H^{(l-1)}W_Q, K^{(l)} = H^{(l-1)}W_K, V^{(l)} = H^{(l-1)}W^V$$

Where $W_Q, W_K, W^V \in R^{d \times d_k}$ is the learnable parameter matrix, and H is the dimension of each head.

Next, the self-attention mechanism calculates the dependencies between positions, specifically:

$$Attention(Q,K,V) = softmax(\frac{QK^T}{\sqrt{d_k}})V$$

This mechanism strengthens the model's attention to key temporal relationships by scaling dot products and normalizing operations. Finally, in the multi-head attention mechanism, the attention results on multiple subspaces are concatenated and linearly transformed, expressed as:

$$MultiHead(Q,K,V) = Concat(head_1,...,head_h)W_O$$

Where h represents the number of heads and $W_O \in R^{hd_k \times d}$ is the output mapping matrix.

Through the above structure, the Transformer can globally model the dependencies between time points and deeply model the implicit patterns in high-dimensional monitoring signals. This module not only can perceive long-distance dependencies, but also provides a unified expression basis for subsequent multi-scale information fusion and abnormal pattern recognition. To adapt to the characteristics of long-sequence high-frequency data in actual cloud platforms, we also introduced a multi-scale perception mechanism in the subsequent part to further enhance the representation ability of the model.

*B. Multi-scale fusion*

To further enhance the Transformer architecture's ability to model multi-granularity time series features, this study incorporates a multi-scale fusion mechanism designed to improve the expressiveness of the model when applied to dynamic cloud service monitoring data. The central concept of multi-scale modeling lies in capturing complementary information from multiple temporal perspectives, which involves analyzing data through varying time windows, abstraction hierarchies, and receptive field ranges. This enables the model to simultaneously capture short-term fluctuations, such as sudden performance shifts, and long-term behavioral trends that develop over extended periods. To realize this, we construct several parallel processing paths on top of the original input sequence. Each path employs a distinct downsampling strategy to obtain time series representations at different temporal resolutions, allowing the model to capture diverse features across multiple scales. These multi-resolution sequences are then integrated to form a comprehensive understanding of the input dynamics. The complete architecture of the model is illustrated in Figure 2.

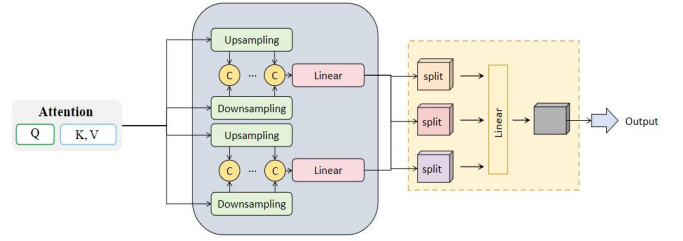

Figure 2. Multi-scale fusion model architecture

Define the original feature sequence as $X \in R^{T \times d}$, and construct the downsampled representation on the sth scale through the scaling function $\phi$:

After completing the multi-scale sequence construction, the subsequences at each scale will pass through an independent Transformer encoding module to model the temporal dependency features of the scale and obtain multiple scale feature representations $H(s) \in R^{T_s \times d}$. To unify the semantic space of these different-scale features, we introduce a scale alignment module to remap them to the same length and dimension, and use a linear transformation function $f_s(\cdot)$ to project each scale:

$$\widetilde{H}^{(s)} = f_s(H^{(s)}), \widetilde{H}^{(s)} \in R^{T \times d}$$

After the alignment, the features are sent to the fusion module for integration to preserve the semantic clues at each scale. We use the attention weighting mechanism for fusion, dynamically learn the contribution weight of each scale to the final representation, and define the attention weight as:

$$a^{(s)} = \frac{\exp(W^T \cdot \tanh(W \cdot \widetilde{H}^{(s)}))}{\sum_{j=1}^{S} \exp(w^T \cdot \tanh(W \cdot \widetilde{H}^{(j)}))}$$

Where $w \in R^d, W \in R^{d \times d}$ is a learnable parameter and $a^{(s)}$ represents the attention coefficient of the sth scale.

The final fusion representation $H_{fusion}$ is the weighted sum of features at each scale, expressed as follows:

$$H_{fusion} = \sum_{s=1}^{S} a^{(s)} \cdot \widetilde{H}^{(s)}$$

This fused representation is fed into subsequent modules as the final unified feature expression. By introducing multi-scale paths, feature alignment, and attention fusion strategies, the model can take into account both local detail changes and global trend modeling, significantly expanding the expression space for complex dynamic structures based on the original Transformer architecture, and forming a multi-granular perception capability for abnormal patterns of cloud services.

## IV. DATASET AND EXPERIMENTAL SETUP

### A. Dataset

This study employs the Alibaba Cluster Trace 2018 dataset, which captures operational traces from a large-scale cloud computing platform. Spanning eight days and covering over 4,000 servers with millions of task instances, the dataset includes job scheduling logs, container allocations, resource utilization metrics (CPU, memory, disk I/O), and annotated anomaly events such as service interruptions, resource surges, and latency drifts. Its scale, heterogeneity, and fidelity make it a widely recognized benchmark for performance modeling and anomaly detection. To ensure usability, the raw data is preprocessed through normalization, missing-value imputation, label refinement, and time-series slicing, producing a unified multidimensional format that preserves structural dependencies while enabling effective anomaly signal extraction.

### B. Experimental setup

This study implements the full model training and evaluation pipeline on a deep learning platform based on the PyTorch framework. All experiments are conducted on a server equipped with an NVIDIA A100 GPU. The system environment is Ubuntu 20.04, and the CUDA version is 11.8. The input to the model consists of standardized multidimensional time series data. A fixed window slicing method is used to construct training samples. The window size is set to 60 with a stride of 1. During training, the AdamW optimizer is used with an initial learning rate of 1e-4. A cosine annealing scheduler adjusts the learning rate. The batch size is 64, and the number of training epochs is 100. An early stopping mechanism is applied to prevent overfitting.

To evaluate the generalization performance of the model under different anomaly scenarios, the dataset is split into training and testing sets with a ratio of 8 2. The distribution of anomaly proportions is kept consistent. All models are trained multiple times independently under the same data split, hyperparameter settings, and random seed to ensure reproducibility and stability. For fair comparison, all baseline models use the same input feature structure and normalization method. The evaluation metrics include precision, recall, F1-score, and AUC. This experimental setup ensures the rigor of the results and the effectiveness of the comparative analysis.

## V. EXPERIMENTAL RESULTS

### A. Comparative experimental results

This paper first gives the results of the comparative test, and the experimental results are shown in Table 1.

Table 1. Comparative experimental results

| Method | Precision | Recall | AUC | F1-Score |
|---|---|---|---|---|
| ResNet[49] | 0.812 | 0.765 | 0.874 | 0.788 |
| VGG[50] | 0.826 | 0.752 | 0.861 | 0.787 |
| LSTM[51] | 0.841 | 0.809 | 0.888 | 0.824 |
| GRU[52] | 0.847 | 0.821 | 0.893 | 0.834 |
| Transformer[53] | 0.868 | 0.844 | 0.912 | 0.856 |
| Transformer+CNN[54] | 0.881 | 0.862 | 0.923 | 0.871 |
| Ours | 0.902 | 0.887 | 0.941 | 0.894 |

Experimental comparisons demonstrate that convolutional models (ResNet, VGG) exhibit weak anomaly detection capability, with low precision and recall arising from their inability to capture temporal dependencies and multidimensional feature interactions. Recurrent architectures (LSTM, GRU) achieve modest improvements in recall and AUC by modeling sequential patterns, yet remain limited by memory constraints and gradient instability. The Transformer baseline surpasses these methods across all metrics, while the Transformer–CNN hybrid further improves performance, attaining an F1-score of 0.871 through the integration of global attention and local convolutional refinement. The proposed approach achieves the highest effectiveness, with recall of 0.887 and AUC of 0.941, confirming its robustness, adaptability, and suitability for anomaly detection in dynamic cloud service environments.

### B. Hyperparameter sensitivity experiment results

Furthermore, this paper gives the results of the hyperparameter sensitivity experiment. First, the hyperparameter sensitivity experiment of the learning rate is given, and the experimental results are shown in Table 2.

Table 2. Hyperparameter sensitivity experiment results(Learning Rate)

| Learning Rate | Precision | Recall | AUC | F1-Score |
|---|---|---|---|---|
| 1e-3 | 0.861 | 0.823 | 0.902 | 0.841 |
| 3e-4 | 0.884 | 0.867 | 0.928 | 0.875 |
| 2e-4 | 0.894 | 0.876 | 0.934 | 0.884 |
| 1e-4 | 0.902 | 0.887 | 0.941 | 0.894 |

The results in Table 2 demonstrate that the learning rate exerts a substantial influence on anomaly detection performance in cloud service environments. A rate of 1e-3 leads to unstable training with low precision and recall, as overly large step sizes cause the model to overshoot the optimal solution region, resulting in degraded detection capability. Performance improves at 3e-4 and 2e-4, where slower updates enhance stability, enabling more accurate capture of local features and anomaly distributions, thereby increasing recall and AUC. The best outcomes occur at 1e-4, with the model achieving an AUC of 0.941 and an F1-score of 0.894, reflecting an effective balance between convergence precision and training efficiency. At this rate, the Transformer-based architecture with multiscale fusion becomes more sensitive to sudden behavioral changes while maintaining a low false positive rate, underscoring its practical value for robust anomaly detection in highly dynamic cloud systems. Overall, these findings confirm that the learning rate is a critical hyperparameter for optimizing temporal structure perception and multi-scale feature modeling, and systematic sensitivity analysis is necessary to ensure stable and generalizable performance. Additional experimental results on different optimizers are reported in Table 3.

Table 3. Hyperparameter sensitivity experiment results(Optimizer)

| Optimizer | Precision | Recall | AUC | F1-Score |
|---|---|---|---|---|
| AdaGrad | 0.864 | 0.842 | 0.910 | 0.853 |
| SGD | 0.848 | 0.819 | 0.898 | 0.833 |
| Adam | 0.887 | 0.869 | 0.931 | 0.878 |
| AdamW | 0.902 | 0.887 | 0.941 | 0.894 |

The comparison shows that optimizer choice strongly affects anomaly detection performance. SGD performs weakest due to slow convergence and local minima, while AdaGrad improves recall but suffers from diminishing updates. Adam achieves stronger stability and accuracy, with an AUC of 0.931 and F1-score of 0.878, benefiting from its adaptive moment estimates. The best results come from AdamW, which reaches an F1-score of 0.894 by combining Adam's efficiency with weight decay for better generalization and reduced overfitting, confirming the importance of selecting advanced optimizers for robust cloud anomaly detection.

### C. Visualizing Experimental Results

In the visualization analysis stage, this paper first gives an evaluation of the sensitivity of the activation function selection to the detection effect, and the experimental results are shown in Figure 3.

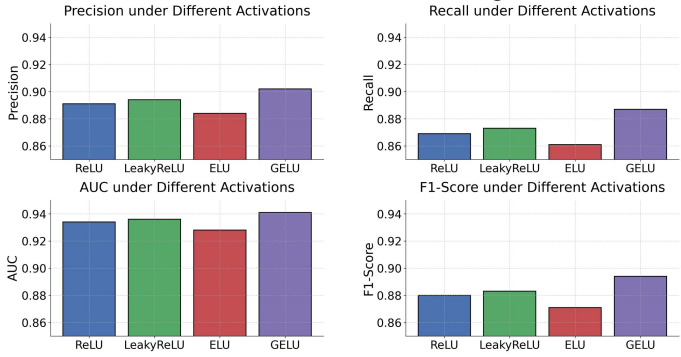

Figure 3. Sensitivity evaluation of activation function selection on detection effect

Figure 3 shows that activation function choice strongly affects anomaly detection, with GELU yielding the highest precision, recall, and F1-score. Its smooth, adaptive mapping enables robust modeling of complex anomaly patterns, unlike ReLU and LeakyReLU, whose hard cutoffs cause gradient loss on multiscale features. ELU performs slightly better than ReLU but remains inferior, as negative saturation cannot fully capture temporal dependencies. Overall, GELU's higher-order differentiability provides stronger representation, confirming the critical role of activation functions in Transformer-based anomaly detection for dynamic cloud environments. The effect of noise injection intensity is further evaluated in Figure 4.

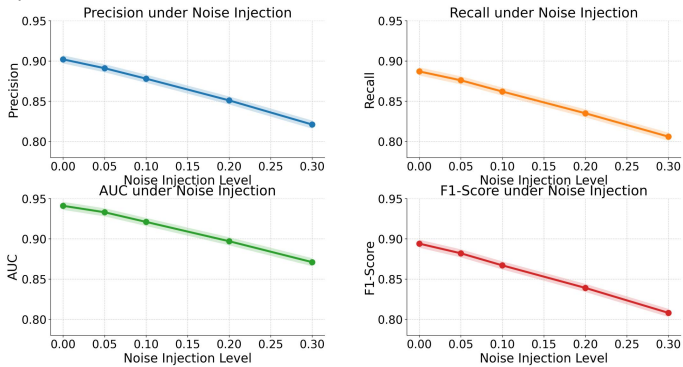

Figure 4. The impact of noise injection intensity on the robustness of detection models

The results in Figure 4 show that increasing noise injection consistently degrades anomaly detection performance, with recall and F1-score most affected, reflecting weakened stability and a higher risk of missed detections in cloud environments. Precision declines more moderately, suggesting boundary shifts under perturbations, while the drop in AUC indicates reduced overall discriminative ability despite the use of multiscale fusion. These findings confirm that although the model performs well in low-noise settings, it faces robustness limitations under noisy conditions, highlighting the need for perturbation-aware learning or denoising strategies. The corresponding loss function trend is illustrated in Figure 5.

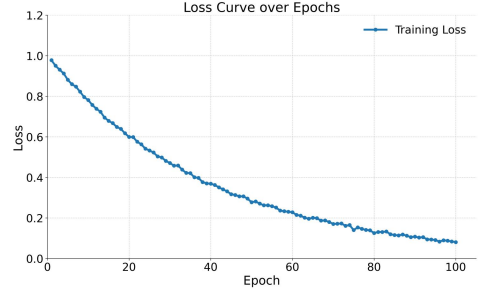

Figure 5. Loss function changes with epoch

As shown in Figure 5, the loss steadily decreases until epoch 40 and then enters a slower, stable convergence phase through epoch 80, indicating that the model has saturated in learning multiscale and dynamic temporal structures while refining subtle anomaly patterns. Beyond epoch 80, the loss remains low and stable without oscillation, suggesting strong generalization and no overfitting. This behavior confirms that, with the multiscale fusion mechanism and improved Transformer encoding, the model effectively captures cross-scale temporal dependencies and boundary features, demonstrating both sensitivity to anomalies and robustness in representation. The smooth decline of the loss further validates the soundness of the architecture and hyperparameter settings, underscoring the model's trainability and adaptability for real cloud service anomaly detection. The sensitivity experiment on changes in anomaly proportion is additionally reported in Figure 6.

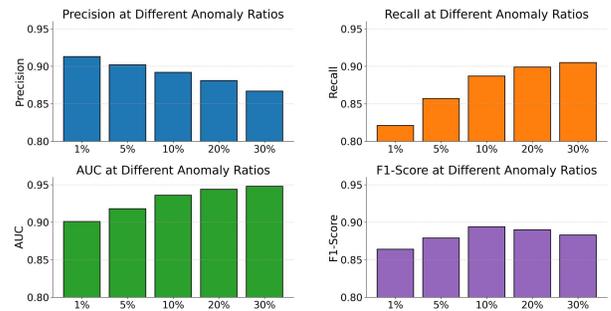

Figure 6. Sensitivity experiment of abnormal ratio change on model recognition ability

The results in Figure 6 demonstrate that anomaly proportion strongly influences detection performance, particularly in the trade-off between precision and recall. At low anomaly ratios (1%–5%), the model achieves high precision, indicating effective suppression of false positives, but recall remains low as rare anomalies are often missed—a scenario consistent with real-world cloud services. As the anomaly proportion increases, recall improves significantly, while precision declines slightly, reflecting a relaxed decision boundary that favors broader coverage at the expense of accuracy. The AUC shows a steady upward trend, confirming enhanced discriminative ability and clearer classification boundaries under higher anomaly densities, while the F1-score peaks around 10%, suggesting this ratio achieves the best balance between precision and recall. These findings highlight that the proposed multiscale fusion framework maintains strong adaptability across varying anomaly distributions and underscore the importance of carefully controlling anomaly ratios in training to build robust and practical cloud service anomaly detection systems.

## VI. Conclusion

This study addresses key challenges in anomaly detection within cloud service environments by proposing a detection algorithm based on the Transformer architecture with integrated multiscale perception mechanisms. The method effectively resolves limitations in temporal modeling and the difficulty of expressing multigranularity anomaly features in high-dimensional monitoring data. By introducing a self-attention mechanism and scale-aware hierarchical representation, the model captures both long-range temporal dependencies and local fluctuations.

In terms of model design, this study introduces a structural innovation that enables multipath and multiscale encoding of input features. It integrates semantic representations from different scales and applies an attention-weighted fusion module. Without relying on additional label information, the model improves its responsiveness to anomalous regions. To address real-world issues such as sparse anomalies and strong noise interference, the input modeling and training mechanism incorporates enhancement strategies and robustness evaluation methods. These components improve the model's generalization and fault tolerance in industrial environments, demonstrating strong potential for practical deployment.

A series of sensitivity experiments and comparative evaluations analyzes the model's performance under variations in learning rate, optimizer, noise perturbation, and anomaly ratio. The results validate the stability of the architecture and the rationality of the parameter settings. The proposed method consistently maintains strong detection performance across different data configurations and anomaly scenarios. This highlights its adaptability and scalability in heterogeneous, multimodal, and dynamically changing cloud computing environments. The study provides both an algorithmic foundation and a practical path for building next-generation intelligent operations systems.

While the proposed multiscale Transformer achieves strong accuracy, the self-attention mechanism scales quadratically with the sequence length $L$ in computation and memory, which can be costly for long windows and high-frequency telemetry; lightweight or sparse attention variants may help reduce latency in deployment. In addition, our experiments are conducted on the Alibaba Cluster Trace 2018, so cross-environment generalization to other clouds and workload mixes may degrade under distribution shift. Evaluating on diverse cloud-oriented benchmarks—e.g., Exathlon, which contains Spark-cluster anomalies such as resource contention and process failures—would help stress-test generalization and guide domain-adaptation or continual-learning strategies [55].

Future research may explore directions such as unsupervised pretraining, incremental learning, and graph-based modeling. These approaches could improve data efficiency under label-scarce conditions. In addition, achieving fast response through lightweight modeling and edge computing architectures in real-world distributed systems is another promising direction. Overall, the proposed multiscale Transformer-based anomaly detection framework expands the technical frontier of intelligent cloud monitoring. It also provides a generalizable solution for temporal modeling and high-dimensional feature fusion tasks, offering significant theoretical and practical value.

## VII. Use of AI

I exclusively used AI tools to assist with grammar and wording, while all the core concepts, analysis, and writing were my original work.